# Modélisation d'une analyse pragma-linguistique d'un forum de discussion

Nada Matta (1)
nada.matta@utt.fr

Karima Sidoumou (1)
karima.sidoumou@utt.fr

Goritsa Ninova (1)
goritsa.ninova@utt.fr

Hassan Atifi (1)
hassan.atifi@utt.fr

*(1) Université de Technologie de Troyes-Institut Charles Delaunay, 12 rue Marie Curie BP. 2060 10010 Troyes Cedex.*

***RÉSUMÉ****. Les chercheurs en pragma-linguistique étudient les interactions afin de comprendre les mécanismes qui régissent ces interactions. Ils ont développé des techniques et des critères permettant de guider l'analyse de ces interactions. Nous avons modélisé la démarche suivie d'un chercheur en pragma-linguistique pour analyser des messages d'entraide (extraits du forum Doctissimo) afin de mettre en avant le choix et l'exploitation de ces critères. Notre objectif est de fournir des guides pour analyser et comprendre les interactions dans un forum de discussion afin d'aider par exemple, à la définition des tags et d'autres types de métadonnées facilitant la compréhension des interactions.*
***MOTS-CLÉS :*** *Ingénierie des connaissances, Analyse conversationnelle, Pragma-linguistique, forums de discussion d'entraide.*

***ABSTRACT.****. We present in this paper, a modelling of an expertise in pragmatics. We follow knowledge engineering techniques and observe the expert when he analyses a social discussion forum. Then a number of models are defined. These models emphasises the process followed by the expert and a number of criteria used in his analysis. Results can be used as guides that help to understand and annotate discussion forum. We aim at modelling other pragmatics analysis in order to complete the base of guides; criteria, process, etc. of discussion analysis*

## 1. Introduction

L'objectif de cette étude est de mettre en avant certaines techniques suivies en pragma-linguistique pour analyser des forums de discussion. Nous avons en effet observé un chercheur en pragma-linguistique dans son travail et nous avons modélisé les différentes étapes qu'il suit pour définir des scripts types des interactions dans un forum de discussion. Les modèles ainsi définis fourniront des éléments pour définir des guides d'analyse et de compréhension des interactions médiatisées. La pragmatique des interactions ou analyse des interactions est une nouvelle démarche pluridisciplinaire, une synthèse de disciplines diverses qui étudient le langage en situation, principalement [Atifi et al.2006]:

- La nouvelle communication : orientation transdisciplinaire anglo-saxonne qui s'est développée dans les sciences psychosociales : ethnographie de la communication, ethnométhodologie, école de Palo-Alto…

- La pragmatique linguistique : née avec les travaux d'Austin et Searle et qui a marqué une rupture radicale avec la linguistique Chomskyenne [Searle, 1979]. [Austin, 1970].

La naissance de la pragmatique des interactions est récente. Elle est le résultat d'un double mouvement qui a touché au début des années 1960 les sciences psychosociales, et à partir des années 1970 la linguistique :

- L'éclatement progressif des frontières qui séparaient traditionnellement les disciplines traitant du langage en contexte (anthropologie, sociologie, psychologie, ...). Ces chercheurs américains ont montré la faiblesse des disciplines autonomes pour traiter correctement du langage en contexte. " En effet, les chercheurs en sciences sociales qui posent des questions pertinentes n'ont généralement pas la formation et les connaissances nécessaires pour traiter l'aspect linguistique du problème ; seulement, la linguistique, qui se trouve au cœur de l'étude du langage, s'est attachée presque exclusivement à l'analyse de la structure du langage en tant que code référentiel, négligeant sa signification, sa diversité et son utilisation sociales." [Hymes, 2003].
- L'évolution de la linguistique traditionnelle vers une nouvelle linguistique orale, pragmatique et communicative. Cette mutation concerne en priorité deux dimensions du langage : sa vocation communicative d'échange et de dialogue et sa nature comportementale et pragmatique. La pragmatique des interactions valorise une perspective sociolinguistique de l'objet étudié en posant les questions suivantes : Qui parle? A qui? De quoi? Quand? Où? Comment? Et Pourquoi? Cette démarche valorise une méthodologie ethnographique basée sur l'observation, la description et l'interprétation des échanges. L'une des formes de la pragmatique des interactions est l'analyse des conversations. Il s'agit d'étudier minutieusement ce qui se produit dans les interactions verbales en s'intéressant spécifiquement à l'organisation séquentielle des tours de parole.[ Kerbrat-Orecchioni, 1993 ].

Nous présentons dans cet article les résultats de la modélisation de l'analyse d'un forum de discussion à dimension sociale.

## 2. Cadre de l'analyse

Le chercheur en pragma-linguistique analysait des extraits du forum Doctissimo (http://forum.doctissimo.fr/) qui correspondaient à des messages d'entraide. Cette analyse était réalisée en collaboration avec un chercheur en psychologie cognitive. L'objectif du chercheur en pragma-linguistique est de coroborer son analyse avec des théories psycho-sociales sur les échanges d'entraide. Il a comme objectifs principaux d'une part de valider les catégories psycho-sociales sur les échanges d'entraide et d'autre part, de définir des scripts type pour chaque catégorie [Gauducheau, Marcoccia, 2007].

Les extraits d'échanges correspondaient essentiellement à une requête et des discussions relatives à cette requête. Comme exemple de ces échanges, nous pouvons citer par exemple : des échanges sur la dépendance aux produits stupéfiants ou des échanges à propos des allergies au lait, etc.

| **Discussions sur l'indépendance aux produits stupéfiants :** | |
|---|---|
| kikou, | Voilà je me lance enfin à ouvrir mon premier post. Ca fait des semaines que j'erre sur le forum, poste des messages par-ci, par-là, mais je n'arrive pas à me poser. Je ne connais encore que très peu de personnes, donc je vais faire une p'tite présentation : J'ai 33 ans et je suis .... Après avoir consommé pas mal de produits, j'ai fini par trouvé mon bonheur (et mon malheur aussi) dans l'héro... à laquelle je suis dépendante maintenant depuis 8 ans. …..<br><br>Je suis aussi dépressive et hyper-angoissée. Quand j'arrête, je prends conscience de la réalité et ca m'est insupportable.<br>Je ne demande rien, je me sens seule avec mes problèmes. Vous pouvez venir ici pour me parler, me raconter votre histoire ou juste pour dire ce que vous avez sur le coeur... |
| Orchidée | Je ne suis pas très loquace aujourd'hui, mais je connais bien le problème de l'héro. Je suis encore loin de m'en être "débarrassée", bref ... Si j'ai un peu de courage plus tard, lolll. |
| Diana | Bonjour, désolée, je ne connais pas ton problème, mais si c'est un appui ou une écoute que tu cherches, je t'ai lue. Je te souhaite bon courage, pour moi c'est l'alcool et le shit (en conso excessive et "famélique" ). |
| **Discussions sur les allergies au lait :** | |
| cartoon181 | Salut à tous, Je m'interroge sur une éventuelle intolérance aux produits laitiers ( quel composant exactement, je ne sais pas ) chez moi. Je m'explique : J'ai longtemps remis en cause mon alimentation dans ma légère acné qui persistante malgré mes 22 ans. Je ne mange pourtant pas très gras, mais je me tappe souvent des éruptions cutanées sur le front, les tempes, qui sortent de je ne sais ou, sans raison précise ... |
| go | C'est difficile à affirmer que les produits laitiers provoquent de l'acné systématiquement, mais ... cela peut en induire chez certaines personnes plus facilement…. |
| joely | Bonjour, je suis a peu près persuadé que le lait en particulier de vache, est mauvais pour la santé. Une idée reçue, répandue par la puissance des lobbys pousse à croire qu'il apporte du calcium,… |

**Figure 1.** *Extraits des échanges étudiés.*

Les catégories psycho-sociales utilisées dans cette analyse sont : demande de soutien émotionnel, demande de partage d'expériences, demande d'évaluation, demande de soutien informationnel et de conseil. Ces catégories sont extraites d'une étude sur les formes du soutien social [Gauducheau, Marcoccia, 2007].

## 3. Observation de l'analyse

Nous avons utilisé les techniques de l'ingénierie des connaissances pour analyser et modéliser l'expertise de l'analyse du forum. L'ingénierie des connaissances (IC) fournit essentiellement une démarche d'analyse et de modélisation d'une résolution de problèmes [Charlet, 2003], [Aussenac et al, 1996]. Des guides méthodologiques sont fournies dans cette discipline. Ils permettent de recueillir une expertise et de la modéliser afin de mettre en avant le « pourquoi », le « quoi » et le « comment » d'un comportement.

Dans notre cas, nous avons suivi la démarche de recueil et de modélisation de l'ingénierie des connaissances. Nous avons observé l'activité d'analyse effectuée par l'expert en pragma-linguistique selon un scénario bien établi. En effet, l'ingénierie des connaissances emprunte des techniques de l'ergonomie pour observer une activité. Il s'agit d'abord de bien identifier l'objectif de la tâche à observer, d'effectuer une série d'observations, d'analyser les observations afin de définir une série de schémas comportementaux, et enfin de les valider avec l'expert en le poussant d'expliquer le pourquoi de son comportement.

Nous avons alterné les observations avec les entretiens de validation et d'explicitation afin de comprendre au fur et à mesure les raisons du comportement de l'expert à chaque étape d'analyse.

## 4. Modélisation de l'analyse

Le raisonnement que suit le chercheur expert en pragma-linguistique relève plutôt de l'abduction. En fait, il analyse un échantillon de messages relevant de requêtes d'entraide, il formule un certain nombre d'hypothèses sous forme de scripts d'analyse et enfin, il valide ces hypothèses sur d'autres scripts (Figure 2. ).

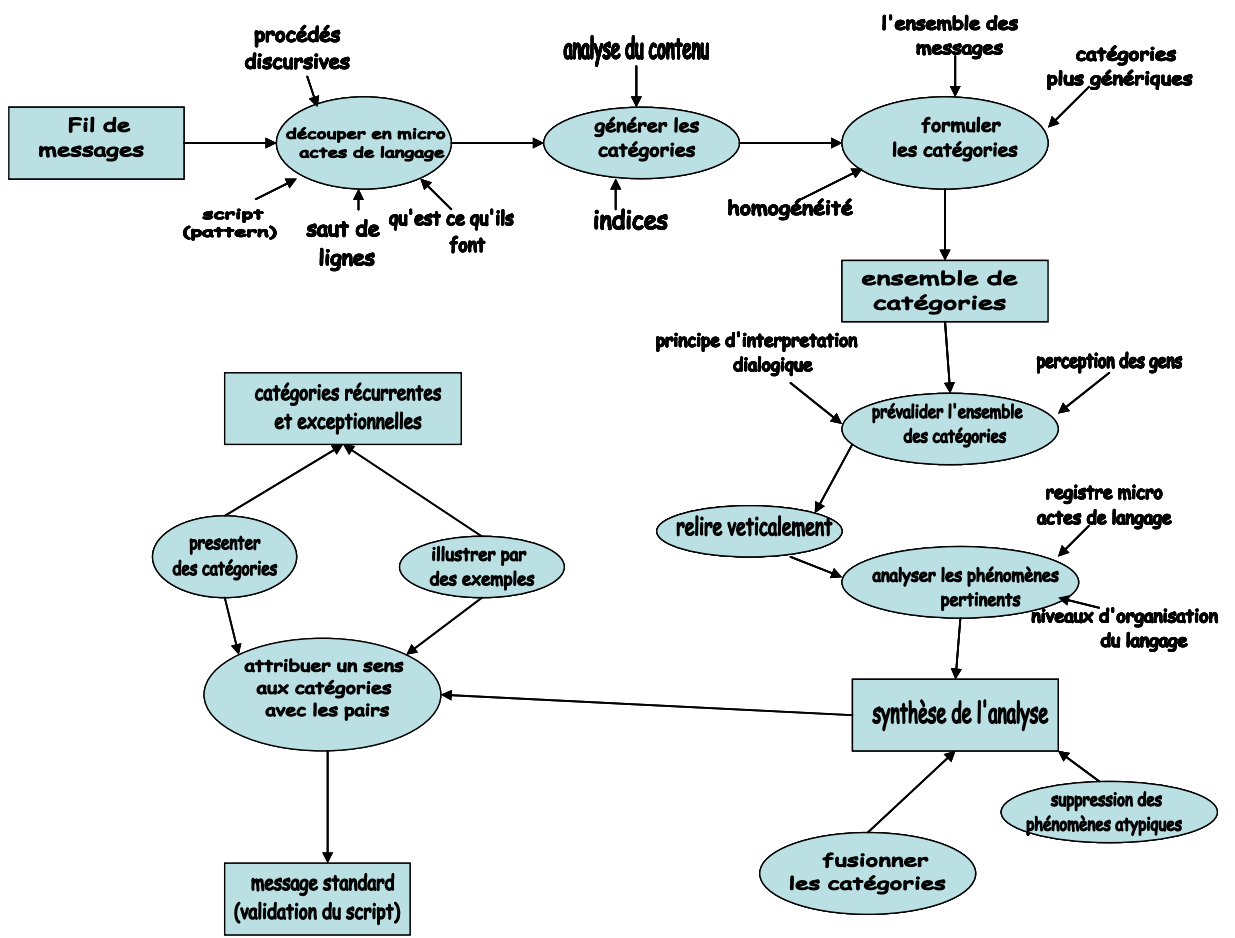

**Figure 2.**   *Analyse des interactions.*

Si on observe de plus près son analyse pour identifier des scripts, nous pouvons identifier qu'il procède par une série d'évaluations en se basant sur des théories pragma-linguistiques [Kerbrat-Orecchioni, 1998], [Hymes, 2003]. Ces théories lui servent de critères dans son analyse. Le modèle générique d'évaluation de CommonKADS [Breuker et al, 1984] défini en ingénierie des connaissances, nous a permis d'identifier que l'expert, établit des classes de décisions (qui sont les scripts relatifs à chaque type de requête) en se basant sur des modèles types. En effet, une fois que les interactions sont découpées en fils (requête, réponses et réponses sur réponses), il compare le fil avec les modèles types en utilisant les critères d'analyse (Figure 3. ). Nous notons que cette analyse est faite d'une façon manuelle.

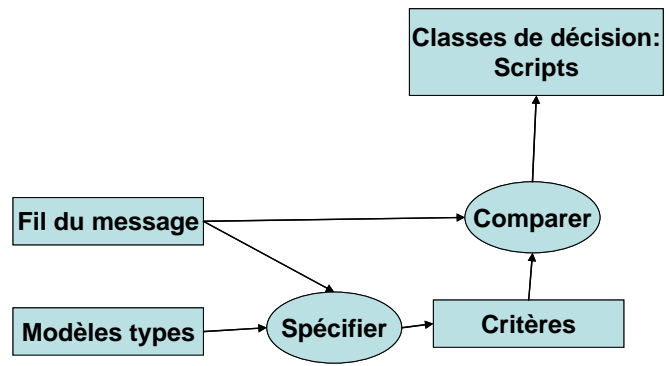

**Figure 3.** *Evaluation d'un fil d'interaction*

Cependant, il cherche en même temps, à identifier des messages non classables dans les catégories identifiées ou qui sont des scripts exceptions. Cela lui permet d'énoncer de nouvelles hypothèses concernant les forums sur les entraides.

### *4.1. Les critères pragma-linguistique utilisés*

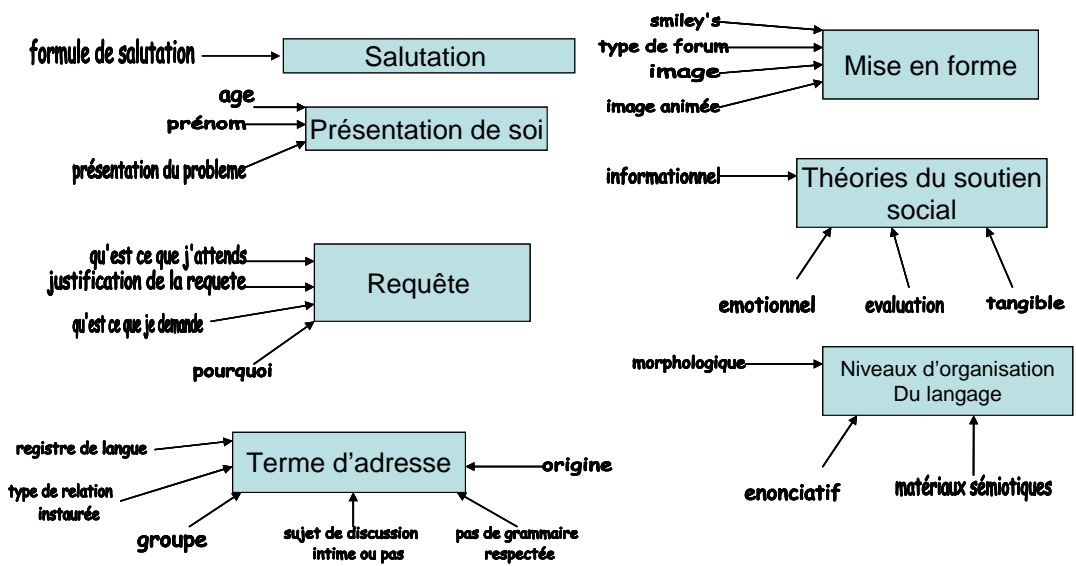

**Figure 4.** *Extrait des critères utilisés dans l'analyse [Hymes, 2003].*

Dans notre modélisation, nous nous sommes intéressés à une analyse fine des messages. En effet, après avoir identifié les grandes lignes de son raisonnement ainsi que les critères qu'il utilise, nous avons observé comment il procède dans son analyse pour découper les interactions et établir les scripts.

Nous présentons en illustrations des traces du « brouillon » de son analyse afin de bien montrer la démarche qu'il suit.

### 4.2. Analyse des échanges

Les interactions sont découpées en fil d'échanges correspondant à chaque requête. Chaque fil comprend une requête et les réponses à cette requête. Le fil est ensuite analysé rapidement pour identifier de quelle catégorie de requête il s'agit : demande de soutien émotionnel, informationnel, demande de conseil, partage d'expériences, etc [Gauducheau, Marcoccia, 2007].

La requête est ensuite analysée de façon à identifier son type et les différents indices comme : les destinataires de la requête, les termes d'adresse, la présentation de l'émetteur, l'explicitation de ce qui est demandé, les salutations, la signature, etc.

Les réactions au message sont ensuite analysées de façon à identifier d'abord une trame de fond comme : les termes d'adresse, les salutations et les signatures, la mise en forme (simleys, etc.), les dictons et les proverbes. Ensuite, d'autres indices sont identifiés selon le contenu des réactions. Par exemple, certains types de réactions ont été identifiés comme : encouragement et compliments, critiques et désaccords, conseils et apports d'informations, évaluation de la situation et demande d'informations complémentaires, évaluation de l'expertise et partage d'expériences proches, etc.

### 4.3. Identification des scripts d'interactions

Une fois les fils d'échanges analysés, l'expert dresse une grille croisée afin d'identifier des scripts par type requête. Cette grille permet d'abord d'identifier les indices récurrents dans chaque type de requête et ensuite de valider les scripts en analysant d'autres fils d'échange afin de valider ses hypothèses.

|  | FIL 1 Requête Demande de soutien émotionnel | FIL 2 requête Demande de partage d'expériences | FIL 3 Requête Demande d'évaluation et de partage d''expériences | FIL 4 Requête Demande de partage d'expériences | FIL 5 Requête Demande de soutien tangible | FIL 6 Requête Demande de soutien informationnel et de conseil | FIL 8 Demande de soutien informationnel et de partage d'expérience |
|---|---|---|---|---|---|---|---|
| Bénéficiaire de la requête | X | X | X | X | X | X | X |
| Salutation d'ouverture | X | X |  | X | X |  | X |
| Terme d'adresse | X | X | X | X |  |  |  |
| Description de l'activité dans le forum | X | X |  | X |  |  |  |
| Identité (âge, origine, etc.) |  | X |  |  |  |  |  |
| Présentation du problème |  | X | X | X |  | X | X |
| Echec dans la résolution du problème |  | X |  | X |  |  |  |
| Etat psychologique | X | X | X | X |  | X |  |
| Etat de santé |  |  | X |  |  |  |  |
| Formulation de la requête | X | X | X | X | X | X | X |
| Description du bénéfice escompté | X |  |  |  |  |  |  |
| Modalités de l'échange et du soutien |  |  |  |  |  |  |  |
| Vœu (contre-don) |  |  |  |  |  |  |  |
| Remerciement par anticipation |  |  |  |  | X |  | X |
| Clôture |  | X |  |  | X |  | X |
| Signature | X |  | X |  |  |  |  |
| Dictons, proverbes, citations |  |  | X |  |  |  |  |
| Mise en forme : smileys, images |  | X |  |  |  |  |  |

Le chercheur constate par exemple, en analysant la grille croisée ci-dessus que les requêtes de type demande de soutien émotionnel se distinguent bien par une présentation de soi bien détaillée et une description des bénéfices escomptés tandis que dans les requêtes de type demande d'autres types de soutien, c'est plutôt le problème qui est plus détaillé en détriment d'une présentation personnelle et du bénéfice. De même, il constate que dans les partages d'expérience, une présentation personnelle et du problème est bien soignée.

## 5. Bilan de la modélisation

L'objectif de la modélisation de cette expertise est de mettre en avant d'une part le type du raisonnement suivi par l'expert observé et d'autre part les indices et critères utilisés dans son raisonnement. Ces indices pourront aider par la suite à l'analyse et la compréhension d'autres interactions du même type, l'entraide.

Nous avons identifié que le raisonnement suivi est une évaluation continue à plusieurs niveaux :

– Découpage des interactions en fil de message : requête, réponses et réponses aux réponses.
– Identification des types de requêtes : dans ce cas, des critères de types psycho-sociaux sont sollicités : demande d'évaluation, demande de soutien informationnel et de conseil, etc.
– Définition des scripts : des indices pragma-linguistiques liés à la structure des messages sont utilisés ; ces indices suivent des scripts bien définis : salutation, présentation de soi, bénéficiaire du message, présentation du problème, explication de la requête ou réponse, signature et clôture.
– Classification des interactions : des indices pragma-linguistiques liés au contenu des messages sont exploités. Nous pouvons citer essentiellement : encouragements et compliments, critiques et désaccords, conseils et apports d'informations, évaluation de la situation et demande d'informations complémentaires, évaluation de l'expertise et partage d'expériences proches.

## 6. Conclusion

Les chercheurs en pragma-linguistique étudient les interactions afin de comprendre les mécanismes qui régissent ces interactions. Ils ont développé des techniques et des critères permettant de guider l'analyse de ces interactions.

Nous avons modélisé l'analyse d'un chercheur en pragma-linguistique d'un forum de discussion afin de mettre en avant le choix et l'exploitation de ces critères. Notre objectif est de fournir des guides pour analyser et comprendre les interactions dans un forum de discussion, par exemple : pour aider à la définition des tags et d'autres types de métadonnées facilitant la compréhension des interactions.

Nous avons étudié l'analyse d'un type précis de messages qui est l'entraide. Les théories mobilisées dans cette analyse relèvent également de la psycho-sociologie. Nous avons montré les scripts spécifiques dédiés à ce type d'interaction. Nous avons modélisé une autre expertise d'analyse pragma-linguistique des interactions concernant l'évaluation d'une activité d'une entreprise. Certains critères comme la présentation personnelle et les salutations sont les mêmes, mais l'analyse du contenu sollicitait plutôt d'autres critères en relation directe avec les actes du langage.

Cette étude n'est qu'à son étape préliminaire et nous envisageons de compléter notre étude et de l'étendre à d'autres types d'interactions afin de déterminer les critères mobilisés ainsi que les scripts correspondant pour chaque type d'interaction.

Nous envisageons de regrouper l'ensemble de ces guides dans un outil support à l'analyse et la compréhension des interactions dans les forums de discussion.

## 7. Références